\documentclass{ecai} 
\usepackage[utf8]{inputenc}
\usepackage{amsmath,amssymb,amsfonts}
\usepackage{algorithmic}
\usepackage{svg}
\usepackage{textcomp}
\usepackage{xcolor}
\usepackage{graphicx,subfig}
\usepackage{url}
\usepackage{svg}
\usepackage{booktabs}
\usepackage{latexsym}
\usepackage{tikz}
\usepackage{pgfplots}
\usepackage{pgfplotstable}
\usepgfplotslibrary{groupplots}

\date{2024}

\begin{document}

\begin{frontmatter}

\paperid{2867}

\title{Measuring User Understanding in Dialogue-based XAI Systems}

\author[A]{\fnms{Dimitry}~\snm{Mindlin}\thanks{Corresponding Author. Email: dimitry.mindlin@uni-bielefeld.de.}}
\author[A]{\fnms{Amelie Sophie}~\snm{Robrecht}}
\author[A]{\fnms{Michael}~\snm{Morasch}}
\author[A]{\fnms{Philipp}~\snm{Cimiano}}

\address[A]{Bielefeld University}

\begin{abstract}
    The field of eXplainable Artificial Intelligence (XAI) is increasingly recognizing the need to personalize and/or interactively adapt the explanation to better reflect users' explanation needs. While dialogue-based approaches to XAI have been proposed recently, the state-of-the-art in XAI is still characterized by what we call one-shot, non-personalized and one-way explanations. In contrast, dialogue-based systems that can adapt explanations through interaction with a user promise to be superior to GUI-based or dashboard explanations as they offer a more intuitive way of requesting information. In general, while interactive XAI systems are often evaluated in terms of user satisfaction, there are limited studies that access user's objective model understanding. This is in particular the case for dialogue-based XAI approaches. In this paper, we close this gap by carrying out controlled experiments within a dialogue framework in which we measure understanding of users in three phases by asking them to simulate the predictions of the model they are learning about. By this, we can quantify the level of (improved) understanding w.r.t. how the model works, comparing the state prior, and after the interaction. We further analyze the data to reveal patterns of how the interaction between groups with high vs. low understanding gain differ. Overall, our work thus contributes to our understanding about the effectiveness of XAI approaches. 
\end{abstract}

\end{frontmatter}

\section{Introduction}

Explainable AI (XAI) is the subfield of AI concerned with developing methods to render predictions or decisions by machine learned models comprehensible for users, organizations, developers etc.~\cite{GuidottiMRTGP19}. Methods range from those that make the general inner workings of machine learning models transparent, such as the meaning of activated neurons in neural networks~\cite{Xu.2019}, to generating explanations of AI behavior for stakeholders using visualizations, knowledge extraction or example-based explanations~\cite{Adadi2018Peeking}.

Most approaches in XAI provide one-shot, non-personalized and one-way explanations in the sense that they deliver a single explanation that is not adapted to the needs of a particular user and that is generated by the explaining system without any possibility for the user to modify the explanation, request further details, ask for an elaboration etc. 
More recently, though, it has been recognized that a single explanation can not meet the needs of all users~\cite{sokolFlach2020one} and that users should be involved in a more participatory way in the explanation process, in particular in the decision of what exactly should be explained and how~\cite{Rohlfing2021}.

One approach that allows users to directly shape the process of generating explanations are dialogue-based XAI systems. Dialogue-based XAI methods promote a dynamic, two-way interaction that allows for explanations to be personalized and contextualized, fostering a co-constructive process where explanations evolve through ongoing user interaction and feedback. This approach aligns more closely with natural human conversational norms and facilitates a deeper, more intuitive understanding of AI systems~\cite{sokol2018glass, miller2019explanation}. By adapting to individual cognitive needs and feedback, dialogue-based methods enhance trust and effectively bridge the gap between complex AI functionalities and user expectations~\cite{hoffman2018explaining}. 

Studies in human-agent interactions, however, present mixed outcomes regarding interactive explanations. Some evidence suggests that while interactive explanations can improve objective understanding, they may also decrease efficiency and user satisfaction due to increased time demands~\cite{robrecht_study_2023}. 
Other research has shown that adaptive explanations do not consistently outperform non-adaptive ones in enhancing the users' understanding~\cite{axelsson_modelling_2019}.
These findings underscore the complexities and varying results across different settings, emphasizing the importance of targeted studies to validate the practical benefits of dialogue-based XAI approaches.

Within conversational XAI, there is a significant lack of studies that quantify users' objective understanding of AI models, with a prevailing focus on collecting subjective feedback~\cite{mindlin2024beyond}. Our research aims to address this gap by using a ``simulation task'' method to objectively evaluate user understanding and the effectiveness of interactive explanatory approaches~\cite{doshi2017towards}. In XAI research, this method involves participants simulating the behavior of the model on unseen examples, ideally complemented by confidence ratings and explanations for their predictions, providing deeper insights into their understanding~\cite{HoffmanMKL23}.

We conduct a study on prolific to investigate whether a dialogue-based setting yields a higher model understanding than a static setting. Our study design incorporates three phases to measure the the user's model understanding after having been exposed to XAI explanations in different conditions. First, in the initial test phase, users undertake the prediction task without exposure to model predictions or explanations, establishing a measure of their intuition ($U_{intuition}$). Next, in the learning phase, participants interact with model predictions and explanations either through a static report (as a baseline condition) or an interactive setting where they can see explanations by asking questions. This phase aims to enhance their model understanding over time, monitored as $U_{time}$. The final phase assesses their model understanding ($U_{model}$). While $U_{model}$ can be seen as a measure of deep enabledness~\cite{buschmeier2023forms}, $U_{time}$ measures how well participants understand and apply the learning of the explanations immediately after receiving them and align their task intuition ($U_{intuition}$) to understanding the model. This structured approach seeks to provide empirical evidence on whether advanced explanatory modalities significantly enhance user's model understanding, thus contributing to the optimal design and implementation of conversational XAI systems.

Our paper offers the following contributions:

\begin{enumerate}
    \item We develop a controlled, dialogue-based experimental framework that allows users to interact with an explaining system over multiple turns by asking pre-defined questions selected from a GUI.
    \item This framework was used to conduct experiments with 200 lay users on a simplified income prediction task, assessing their understanding of model decisions through a simulation task.
    \item We compare the dialogue-based setting to a static setting, showing highly significant results that the dialogue-based approach yields a higher model understanding compared to a static explanation (Mann-Whitney-U-Test; p<0.004).
    \item We analyze and identify common question patterns that users ask to understand model decisions, comparing those who show highest versus lowest model understanding.
\end{enumerate}

\section{Related Work} 

We discuss three lines of work that are relevant for our paper. On the one hand, we discuss approaches that go beyond one-shot explanations. Second, we discuss related work in the field of dialogue-based XAI approaches. Finally, we discuss work related to quantifying the objective understanding of a model by the explainees, that is the receivers of an explanation. 

\subsection{Beyond single Explanations}

Several studies have demonstrated the effectiveness of providing multiple explanations rather than a single explanation.

For example, \citet{baniecki2023grammar} found that providing multiple sequential explanations improved domain specialists' understanding of incorrect model predictions compared to a single explanation. However, the order of these explanations was fixed, not accommodating individual preferences in selecting explanations.

Furthermore, \citet{arora2022explain} extended sentiment classification explanations by letting users alter sentences and see model confidence changes, demonstrating that global cues with feature attributions aid meaningful edits. While up to two explanations are shown, the authors do not allow participants to ask for explanations but provide explanations in a one-shot manner.

Research in Human-Robot Interaction (HRI) has revealed interesting insights into explanations as a communicative process. Based on strategies observed in human-human explanation settings, \citet{RobrechtK23} introduced an adaptive explainer to explain a board game. \citet{axelsson_modelling_2019} have used adaptive explanation strategies to explain art in a gallery, and \citet{buschmeier_attentive_2017} has introduced an agent that is able to do cooperative time scheduling with the user.

Drawing on these findings, it is evident that conversational XAI systems are highly advantageous, facilitating personalized interactions and offering multiple complementary explanations tailored to individual needs and preferences. 

\begin{figure*}[ht!]
  \centering
  \subfloat[Dialogue Explanations]{\includegraphics[width=0.72\textwidth]{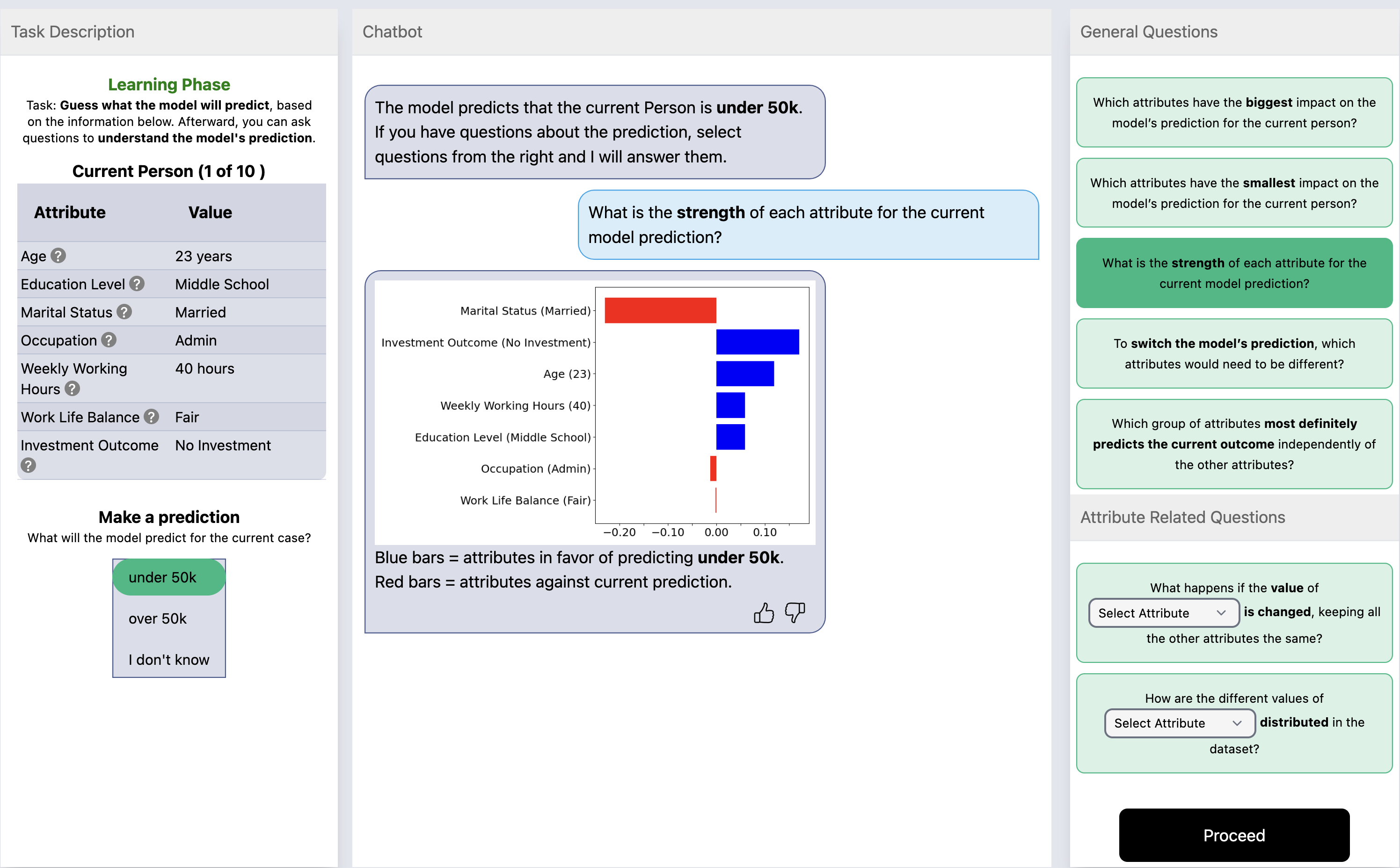}\label{fig:dialogue-ui}}
  \hfill
  \subfloat[Testing Step]{\includegraphics[width=0.25\textwidth]{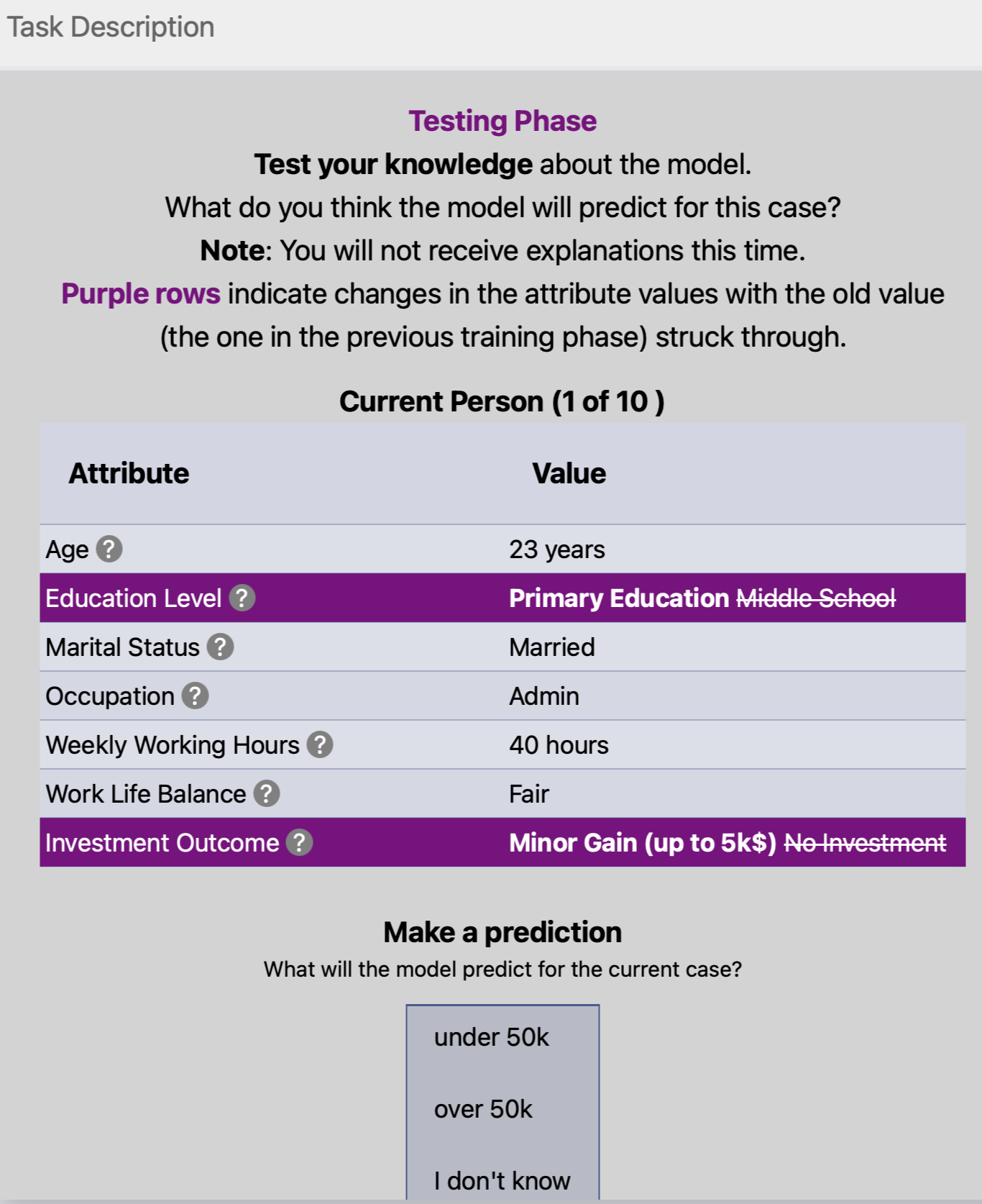}\label{fig:testing-ui}}
  \caption{Experiment UI. a) shows a teaching step in the learning phase of the interactive condition. After a prediction is selected on the right, the Chatbot and Question panel pop up to see the true model prediction and engage in explanations. b) shows a testing instance in the learning step, where the modifications are indicated.}
\vspace{10pt}
  \label{fig:interface}
\end{figure*}

\subsection{Conversational XAI Systems}\label{sec:rel-work-dialog-xai}

Research in conversational XAI so far has focused on tailoring explanations by collecting user feedback. \citet{sokol2018glass} pioneered this with the ``Glass-Box'', gathering feedback at a conference to identify key features for personalized explanations, which improve transparency in machine learning models~\cite{sokolFlach2020one}. \citet{KuzbaB20} developed a conversational model to collect queries from the data science and R communities. While these studies motivate the importance of incorporating user needs to refine explanation systems in XAI, their approach lacked an objective assessment of user understanding improvement, a gap our research addresses.

In the realm of conversational interfaces, both \citet{malandri2023} and \citet{slack2023talktomodel} present innovations that aim to enhance user interaction, such as improved intent understanding and dialogue management, yet do not rigorously assess objective improvement in user understanding and rather rely on qualitative feedback. While their assessment reveals a user preference for the chat interface over traditional dashboards, it does not extend to an objective evaluation of enhanced understanding, highlighting the gap in demonstrating actual understanding improvements in conversational XAI. Similarly, as~\citet{FeldhusWACO023} apply conversational XAI systems to tasks like NLP model understanding, the reliance on qualitative evaluations further underscores the need for more objective assessment methods across the field.

The reviewed studies show advancements in conversational XAI systems with increased personalization and new capabilities. However, the absence of comparable objective evidence on user understanding points to potential risks of improvements in the wrong directions that could distract from increasing understanding and favor aspects such as likability of the systems.

\subsection{Measuring the Impact of multiple XAI Explanations on Understanding Improvement}

The previous section highlighted advancements in dialog systems within XAI, yet underscored the shortfall in rigorous evaluations of user understanding. While there are many studies comparing the influence of one-shot explanations on user understanding (see \cite{wang2022effects} for an overview), there is only a limited amount dedicated to measure the impact of providing multiple explanations and different interface types.

\citet{cheng2019explaining} tested whether an interactive interface that allowed to access model predictions by customized changes would increase model understanding. Understanding was assessed through crafted questions that tested different aspects of understanding like asking about the influence of attribute changes on the model's prediction or a simulation task. However, since the authors did not account for users' intuition, they cannot conclusively attribute the observed understanding improvements to the explanations provided.

\citet{hase2020evaluating} isolate the effect of explanations by including a pre-assessment of the understanding before providing any explanations. Given this, they estimate the change in understanding by measuring the accuracy on simulation tasks, and compare different explanation approaches as well as a composite method that combines multiple explanations. Interestingly, they do not show clear trends that all explanations help to increase understanding. Especially, the composite explanation condition did not yield a significantly higher understanding. Lastly, the explanations were fixed and neither interactive nor selective—two elements we aim to explore in our research.

\citet{baniecki2023grammar} assess model understanding by a different proxy task that involves selecting whether the model's predictions were correct and which of the consecutive explanations was most helpful. These consecutive explanations are similar to a dialog, as they are interrelated and provide a drill-down approach. Again, since the order of these explanations was fixed, the authors did not evaluate the impact of customizing the order and quantity of explanations on understanding.

We identified a single study dedicated to objectively measuring the impact of an explanation interface~\cite{cheng2019explaining}, that missed to isolate the interface's effect on understanding. Other studies that provided multiple explanations did so unidirectionally, limiting the comparability to a dialogue-based setting. We aim to fill this gap and propose an experiment setup to measure model understanding in a conversational XAI setting where we isolate the effect of the interface.

\section{Methods}
This section provides an overview of the experiment framework that we suggest as well as the methods we employ to run the experiment. The code can be found in our github repository\footnote{https://github.com/dimitrymindlin/Measuring-User-Understanding-in-Dialogue-based-XAI-Systems}.

\subsection{User Interface and Dialogue System}

The dialogue system is presented as a web application, depicted in Figure~\ref{fig:interface}. It features an interaction paradigm where users see the attributes of an instance on the upper left, complete with detailed descriptions (available upon hovering on the ?). Users interact by selecting questions on the right; responses appear in the central Chatbot window, which displays the conversation history (see~\ref{fig:dialogue-ui}). In the static condition, all explanations are presented within a report, categorized by topics and organized in the same order as the questions outlined in the interactive condition, rather than in a dialogue window or questions panel. The interface where users are expected to make a prediction is shown in Figure~\ref{fig:testing-ui}. It indicates modifications applied to the learned instance during the learning phase, though this is not shown in the initial and final testing phases.

In developing our dialogue-based XAI system, we chose not to include Natural Language Understanding (NLU) for two primary reasons. First, using a chatbot with predefined questions guarantees clear, error-free interactions by directly mapping queries to responses, though it limits dialogue flexibility. Second, implementing NLU like in TalkToModel~\cite{slack2023talktomodel} would necessitate extensive resources to process around 40,000 phrases for new datasets. This decision simplifies our system and sharpens our focus on evaluating user interaction and the practicality of our experimental design.

\subsection{Experiment Framework}\label{sec:experiment-flow}

\begin{figure}
    \centering
    \includegraphics[width=\columnwidth]{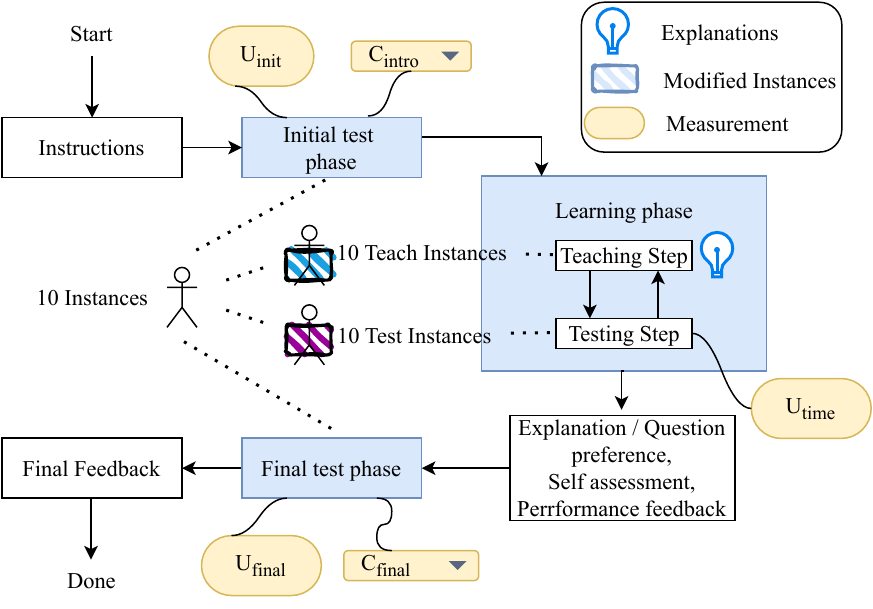}
    \caption{Flow diagram of the study steps: The blue phases require participants to make predictions for displayed instances. The yellow fields indicate assessments of understanding and confidence. While the initial test and final test phases involve the same instances, the instances for the learning phase are modifications of those.}
    \vspace{10pt}
    \label{fig:study-flow}
\end{figure}

Participants begin the experiment by reviewing and agreeing to the study conditions, followed by submitting demographic details and describing their familiarity with machine learning and AI. Afterward, they were shown descriptions and animations of the different tasks for their study condition. The structure of the experiment, depicted in Figure~\ref{fig:study-flow}, consists of three distinct phases: the \textit{initial test phase}, the \textit{learning phase}, and the \textit{final test phase}. Each phase is designed with a specific objective, collectively contributing to the assessment of the participant's understanding.

The initial phase assesses participants' intuition. In this phase, participants make predictions on instances using solely their pre-existing knowledge, without access to model predictions or explanations. This yields a score, $U_{intuition}$, based on correct classifications, that can inform whether the groups have strong differences before engaging with the explanations.

During the learning phase, participants partake in teaching and testing steps. In the teaching step, they first predict outcomes for an instance to enhance engagement through cognitive forcing, a mechanism to delay the display of explanations~\cite{BucincaLGG20}. This prediction is made before they are shown the model's prediction and explanations. The conditions differ in the way the explanations are presented:
\begin{itemize}
    \item \textbf{static condition}: explanations are provided through a written report that lists all the available explanation answers
    \item \textbf{interactive condition}: interface where users can select the question they would like to ask and see the sequence of questions and answers displayed as a chat window (see Figure~\ref{fig:dialogue-ui})
\end{itemize}

The testing step involves predicting outcomes for a modified instance with highlighted changes (See~\ref{fig:testing-ui}), producing a score termed ``understanding over time'', $U_{time}$, based on correct predictions. This cycle promotes reflection and immediate application of new knowledge, enhancing understanding of the model’s decision-making and addressing potential overconfidence, rather than expecting long-term retention~\cite{hase2020evaluating}. After completing the learning phase, participants are asked to rate their subjective understanding and state their preferred types of questions and explanations. Additionally, we inform them of their performance by disclosing the number of test instances they correctly predicted during the phase.

In the final phase, participants are tasked to predict how the model would classify the instances they encountered during the initial test phase, again without the aid of the model’s predictions or explanations. We then count the number of correctly classified instances to calculate the model understanding, $U_{model}$, later via Item Response Theory~\cite{embretson2013item}. This step assesses participants' objective model understanding, facilitating a between-subject comparisons across the two study conditions. This design allows us to distinctly measure and attribute model understanding to either the interactive or static settings.

\begin{table*}[]
\centering
\renewcommand{\arraystretch}{1.3}
\small
\caption{Question-Answer Mapping, Ordered as Presented in Experiment}\label{table:questions}
\begin{tabular}{@{}p{7cm}p{2cm}p{8cm}@{}}
\toprule
Question & Method & Answer \\
\midrule
Which attributes have the biggest impact on the model's prediction for the current person & Lime~\cite{Ribeiro0G16} & "Here are the 3 most important attributes for the current prediction: Investment Outcome is the most important attribute..." \\
Which attributes have the smallest impact on the model's prediction for the current person & Lime~\cite{Ribeiro0G16} & "Work Life Balance is the least important attribute..." \\
What is the strength of each attribute for the current model prediction? & Lime~\cite{Ribeiro0G16} & feature contributions plot \\
To switch the model’s prediction, which attributes would need to be different? & DICE~\cite{MothilalST20} & "Here are possible scenarios that would change the prediction to under 50k: 1. Changing Occupation to Specialized and ..." \\
What is the scope of change permitted to still get the same prediction? & Anchors~\cite{Ribeiro0G18} & "If you keep these conditions: ... the prediction will stay the same." \\
What happens if the value of \textit{Investment Outcome} is changed, keeping all the other attributes the same? & Ceteris Paribus~\cite{kuzba2019pyceterisparibus} & "Changing Investment Outcome to \textit{Major Gain (above 5k\$)} will switch the prediction to over 50k." \\
How are the different values of \textit{Age} distributed in the dataset? & feature statistics & "\textit{Age} ranges from 17 to 90 with a mean of 38,42." or barplot of categories \\
\bottomrule
\end{tabular}
\end{table*}

\subsubsection{Questions and Explanations}

In a literature review, \cite{mindlin2024beyond} highlighted the comprehensive Question Bank by~\citet{liao2020questioning} from conversational XAI studies, which we use as the primary framework for selecting our user queries.

We designed our XAI system using local explanations and feature statistics techniques, drawing on frameworks like TalkToModel~\cite{slack2023talktomodel} and incorporating further question collections from~\citet{liao2020questioning, nguyen2022explaining}, and \citet{malandri2023}. Inspired by the question bank of~\citet{nguyen2022explaining}, we included Anchor explanations~\cite{Ribeiro0G18Anchors} to identify necessary conditions to keep the current prediction. Furthermore, we included Ceteris Paribus~\cite{kuzba2019pyceterisparibus} explanations to demonstrate the impact of feature modifications on the model's decisions, as this question is discussed prominently by~\citet{nguyen2022explaining} as well as~\citet{malandri2023}. We align the answers closely to the work of \citet{nguyen2022explaining} but adapt them slightly for our binary classification setting.

We provide the participants with a range of general and feature-specific questions that explain the model prediction. Our questions are tailored for a general audience and are influenced by a recent study that gathered information through a user survey~\citet{malandri2023}. The final questions and their corresponding answer methods, presented in Figure~\ref{table:questions}, were evaluated for clarity and comprehensiveness in our prestudy, as detailed in Section~\ref{sec:preliminary-study}. While the interactive group was shown explanations based on clicking on the questions, the static group saw all explanations at the same time listed a static report.

\subsection{Measuring Users Model Understanding}

According to \citet{KuleszaBWS15}, understanding is assessed through two mental models: functional and structural. Functional understanding enables operation of a system, while structural understanding provides deeper insights into its workings. Similarly, \citet{buschmeier2023forms} distinguish between enabledness (similar to functional understanding) and comprehension (deeper knowledge of system mechanics). They also introduce a spectrum from shallow to deep for both comprehension and enabledness.

In this study, we define objective understanding as functional understanding, which is generally easier to measure than deep structural knowledge. We focus on assessing measuring how well users can predict a model’s behavior after being exposed to the model prediction and explanations. This is often measured through prediction tasks, where users predict outputs for different instances, reflecting their grasp of critical attributes and system operations~\cite{weld2018intelligible}.

It has been shown that simulatability (prediction tasks) and decision-making in AI systems are influenced by factors such as the users' background knowledge and the accuracy of model predictions and explanations~\cite{Morrison2023Imperfect_Arxiv}. Additionally, the complexity of the domain and task should be appropriately challenging~\cite{hase2020evaluating, robrecht_study_2023}. To manage these variables and ensure a controlled evaluation, several design decisions were made:

Choosing a domain with intuitive features reduces the cognitive load associated with understanding the instances, allowing more time to focus on the explanations. To consider the participants' intuition, we introduce an initial test phase that assesses task performance prior to learning. Next, we focus on explaining the model's decision boundaries without differentiating between correct and incorrect predictions. Recognizing the difficulty in pre-assessing the complexity of prediction tasks, we apply Item Response Theory~\cite{embretson2013item} (one parametric). This theory helps adjust for varying difficulties across prediction instances post experiment to more accurately measure the model understanding. Lastly, unlike previous studies that separate the training and testing phases~\citet{hase2020evaluating}, our approach in part interweaves these phases by following each training instance with a testing instance, before the final test phase. This strategy aims to improve information retention by ensuring application of the learned information, addressing the challenge of participants not retaining information across distinct phases.

\paragraph{Instance selection}

In simulation tasks, our preliminary study (Section~\ref{sec:preliminary-study}) indicates that test instances should not be markedly different from the training ones to maintain task relevance. However, overly similar instances might simplify the task excessively. For the main instances that are selected for the initial and final test, we use a balanced set of 50\% low and high income people with equally distributed values of the most important feature. For the teaching and testing instances, we use a balanced approach by modifying features of the previously selected instances. The test instance is a further modification of the teaching instance, resulting in a closely related instance, ensuring it likely remains within the same decision region. This method, similar to the approach described in \citet{hase2020evaluating}, involves randomly modifying 1 to 3 features of the training instances to create teaching and testing instances. Our approach considers the true feature value distribution and maintains a balanced ratio between modifications that change the model prediction and those that do not.

\section{Experiments}

\begin{figure}[!h]
  \centering
  \includegraphics[width=0.5\textwidth]{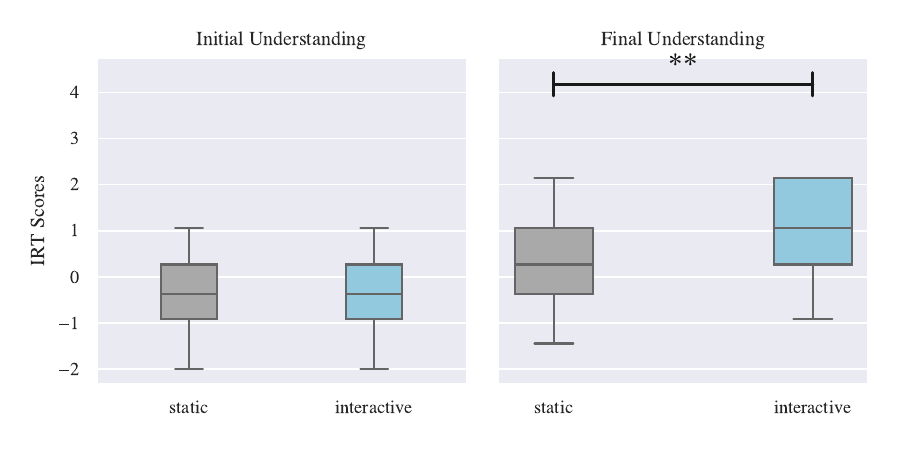}
  \caption{Statistical analysis of difference in intuition and model understanding across conditions. No significant difference in intuition and highly significant difference in model understanding.}
  \label{fig:initial-final-score}
    \vspace{14pt}
\end{figure}

\subsection{Pre-study}\label{sec:preliminary-study}

In preparation for the main study, a pre-study with 80 computer science bachelor students was conducted. Feedback from this pre-study informed refinements to our experimental design, detailed in Section~\ref{sec:experiment-flow}. Three changes in the study design were made based on the pre-study results. The dataset domain should be easily understandable by the participants, as the (1) medical feature names in the Pima Indian Diabetes dataset~\cite{dua2017uci} posed challenges for participants. Some students commented that they found themselves searching online for additional information or resorting to guesswork to make predictions. Furthermore, the (2) exclusively numerical nature of the dataset's features also presented difficulties. Participants struggled to recall and apply information learned during the learning phase to new instances, especially since numerical thresholds varied across instances and were hard to memorize. Therefore, we chose the Adult dataset from the UCI repository~\cite{dua2017uci} for its intuitive, mostly categorical features, simplifying user interaction. Lastly, our initial choice of using (3) randomly selected instances as test cases during the learning phase was ineffective. The high degree of variance between these instances and the teaching instances hindered the application of learned insights. Instead, the main study uses test instances that are slight modifications of the teaching instances to enhance the learning experience.

\subsection{Datasets and Model}

We utilize the \textit{Adult} dataset from the UCI Machine Learning Repository~\cite{dua2017uci}, which comprises 15,682 labeled records of individuals. These records are categorized based on whether the individuals' annual income exceeds \$50,000 or is below this threshold. We use a similar data processing as~\citet{Ribeiro0G18Anchors}, but only include a subset of features as seen in the examples in Figure~\ref{fig:dialogue-ui}. We trained a Random Forest Classifier using the sci-kit-learn library~\cite{pedregosa2011scikit} (ROC-AUC: train: 0.914, test: 0.9).

\subsection{Participant Details}

We ran the study on prolific with 200 participants in April 2023, with an average payment of 10.57£ per hour and median completion time of 26 minutes. The total cost of running the study were 1.428€. Participants were assigned to one of two conditions: interactive or static. To ensure high-quality results, we recruited participants who had an approval rate of at least 99\%, resided in the UK or US, and were native English speakers. To ensure engagement, we implemented three mechanism. First, we incentivised participants by offering additional payments to those who scored in the top 10\% in the final test. Second, we introduced a random variable named "Work Life Balance" into the instances. This variable was consistently presented as the least important, contributing neither to the final prediction nor included in the anchors or counterfactual explanations. It served as an attention or engagement check by filtering out participants who justified their choices in the final test based on this irrelevant attribute. Finally, we limited our analysis to participants who completed the entire study, passed at least one Instructional Manipulation Check (IMC), and did not finish the study in less than two standard deviations below the mean study duration.

After this preprocessing the data, we were left with 107 users of which 54 users were in the static condition and 53 users in the interactive condition. This sample size was deemed sufficient to achieve a power of 0.8 for statistical analysis. The age of participants ranged from 19 to 70 years, with an average age of 40.77 years and a standard deviation of 12.05. Regarding gender, the study included 62 males and 43 females. The self-reported knowledge levels of applications of Machine Learning among respondents are as follows: very low (10), low (44), moderate (43), highly knowledgeable (8), high proficient (2). Given that most respondents rated their knowledge as moderate or lower, where ``moderate'' reflects a basic understanding of AI applications, our study targets lay users rather than experts or professionals in the field. 

\section{Results}

\subsection{Impact of conditions on understanding and Confidence}

To address variations in prediction difficulties across instances, we employed a one-parameter Item Response Theory (IRT) model to assign performance scores to participants. Item difficulties were calculated separately for intuition and model understanding. To obtain both scores and account for the statistical nature of the IRT, we averaged the results over 100 runs and will report the mean and std values for the U statistic and p-values. The intuitive performance across groups was not significantly different (Mann-Whitney U: Mean = 1329.13, Std = 38.826; p: Mean = 0.27, Std = 0.078), and allows us to attribute the model understanding to the condition. We assessed the effectiveness of the two conditions by comparing the model understanding scores, $U_{model}$~(Figure~\ref{fig:initial-final-score}). The boxplots representing $U_{model}$ show that the interactive group had highly significant results and a higher understanding than the static group (U=1002.99\footnote{std=34.52}; p=0.004\footnote{std=0.003}).

We did not observe significant difference in understanding over time, subjective understanding, or confidence in the final predictions.

\subsection{Impact of question selection on understanding}

\begin{figure}
    \centering
    \includegraphics[width=\columnwidth]{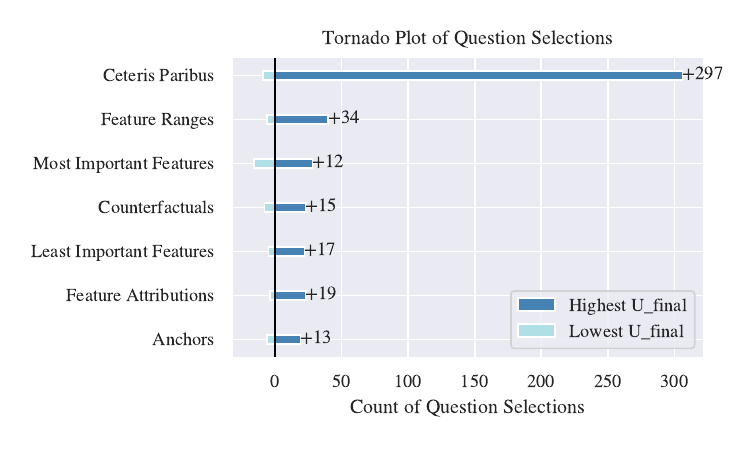}
    \caption{Questions selected for best vs worst understanding improvement users.}
    \label{fig:questions-tornado}
    \vspace{15pt}
\end{figure}

We analyze the questioning behavior between high understanding improvement and low improvement users by dividing interactive users based on their individual improvement in model understanding. This division helps us identify factors contributing to understanding improvement. We define the high performing participants where $U_{improve}>=3$ and worst performing where $U_{improve}<=0$, and after balancing by the amount of the smaller group, resulting in 15 users per group. Figure~\ref{fig:questions-tornado} reveals that the ``high'' group engaged significantly more in the Summarized Ceteris Paribus explanation and somewhat more in Feature Ranges, both of which involve exploring individual features. They also engaged more frequently, and second most often, with Most Important Features to gain an overview of which features are crucial. Interestingly, both groups rated Ceteris Paribus explanations as the least useful in the Question Preferences step.

\paragraph{Exploratory Analysis}

\begin{figure*}
  \centering
    \subfloat[Participant with lowest final understanding]
    {\includegraphics[width=0.45\textwidth]{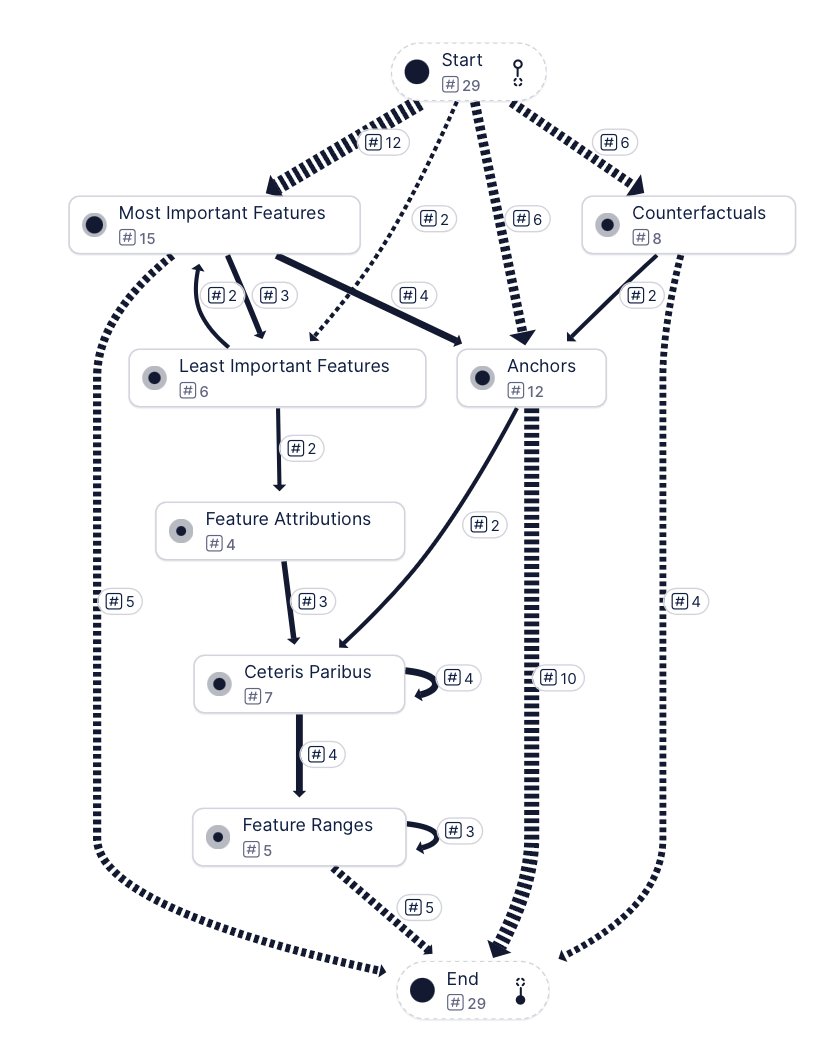}\label{fig:pm-worst}}
  \hfill
  \subfloat[Participant with highest final understanding]{\includegraphics[width=0.45\textwidth]{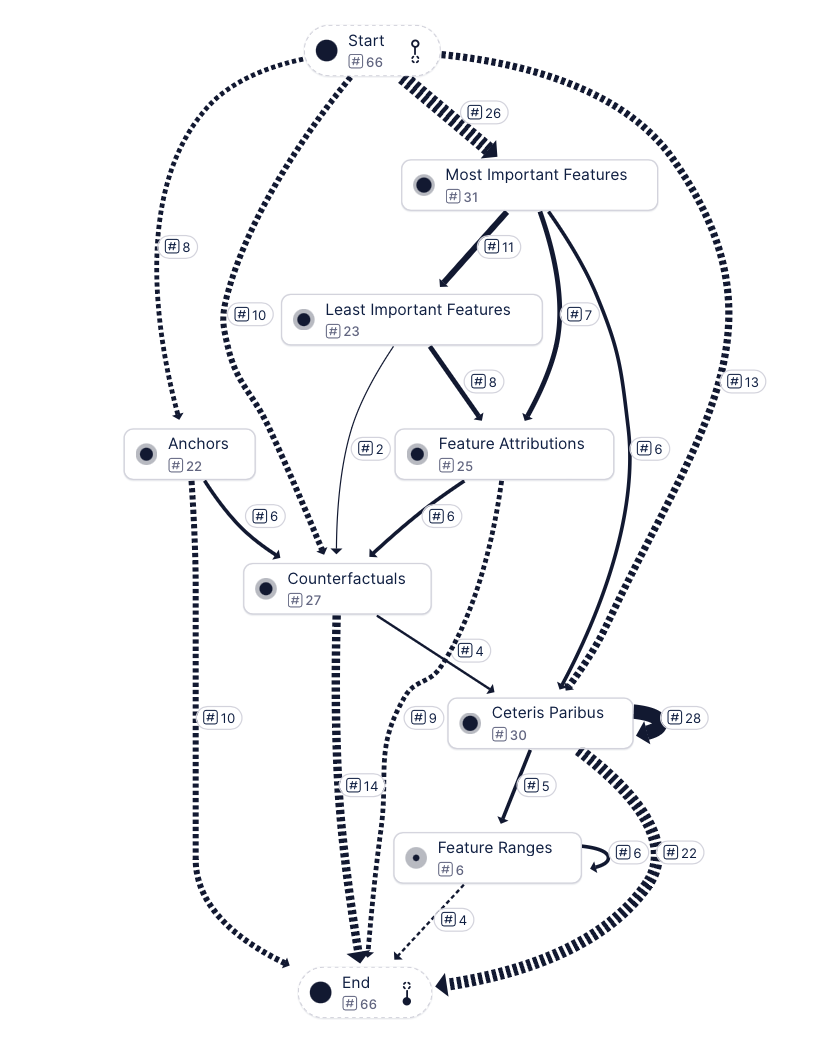}\label{fig:pm-best}}
  \caption{Question Sequences as process mining graphs for participants with lowest and highest final understanding.}
  \label{fig:combined-pm}
  \vspace{10pt}
\end{figure*}

Furthermore, we apply process mining techniques to investigate question sequences, treating each dialogue as a distinct process, as suggested by~\citet{booshehri2024modeling}. Figure~\ref{fig:pm-best} presents the process graphs created using Celonis\footnote{https://www.celonis.com} software for participants who showed the highest improvement, while Figure~\ref{fig:pm-worst} displays those for participants with the least improvement.

Participants with significant improvement typically began by asking about the most important features before delving into Ceteris Paribus explanations. This structured approach suggests that understanding the key features provides a foundation before exploring individual feature impacts. In contrast, the least performing participants more often started directly with Ceteris Paribus explanations, lacking this initial overview from more general questions.

Further, high performers used various explanations in different sequences, deepening their understanding of the model. On the other hand, participants with low improvement followed shorter, more direct sequences before proceeding, potentially missing out on insights.

\paragraph{Correlation Analysis}\label{sec:correlation-analysis}

In our final analysis, we examined the correlations between variables using Pearson correlation coefficients to assess their relationship with the final understanding across groups. We observed a low positive correlation between initial and final understanding (r=0.43) in the static group. Beyond these findings, there were no significant correlations detected.

\section{Discussion}

Our proposed experiment setup allows to elicit user understanding after interacting with a dialogue-based XAI tool. We investigate the relative effectiveness of interactive versus static interactions in helping lay users understand machine learning models. Our results include:

\begin{itemize}
    \item Our three-phase experiment framework effectively measures objective user understanding through prediction tasks comparing different interface designs.
    \item Interactive explanations significantly increase objective model understanding, with no substantial change in subjective understanding reported by users.
    \item Participants who achieved the highest understanding gains explored individual attributes more thoroughly by asking feature-specific questions and, overall, selected more questions.
\end{itemize}

\paragraph{Higher Understanding in Conversational Setting}

The results suggest that interactive groups, which engaged with explanations selectively, achieve higher model understanding. The interactive format likely promotes deeper cognitive processing by requiring participants to actively consider which questions to ask, supporting the co-construction of explanatory dialogues. This implies that interactivity is a critical component in learning environments, particularly when dealing with complex subjects such as AI models.

\paragraph{Analysis of Individuals Understanding Improvement}

The comparison between participants with high versus low model understanding reveals significant differences in both the quantity and types of questions they used to investigate the model decision. The former focused more on feature-specific questions, and did not finish the dialogues after the general questions as often as the latter group, suggesting an effective educational strategy for complex subjects. This finding aligns with the explanation sequence proposed by~\citet{baniecki2023grammar}, which recommends beginning with overviews of global feature importances, followed by a deeper exploration of key features. Higher engagement with questions among participants with higher understanding suggests that accessing more explanations enhances model understanding. These findings support developing XAI systems that promote layered exploration of AI decisions, from general overviews to specific details, while not overwhelming the users with all information at once as in the static condition.

\paragraph{Resulting follow up questions}

Our findings prompt several key questions to further enhance dialogue-based XAI and user understanding: Which aspects of interactive explanations—such as the freedom to explore, or depth of engagement—most effectively enhance understanding? Additionally, what is the optimal sequence for delivering explanations to ensure comprehensive understanding of AI models? Should this sequence be structured and guided, starting from general to specific, or be driven by the explainee's inquiries?

\paragraph{Limitations}

Our study's limitations include our focus and result interpretation for lay users. While it's controlled design, using predefined questions, does not mirror real-world dialogues where users ask varied questions, this yields as a baseline dialogue system. Lastly, the format of the online study introduces some negative effects as user's tendency to quickly finish the studies, which we tried to minimize be excluding fast and inattentive participants.

\section{Conclusion}

Our study provides a systematic analysis of interactive explanatory systems, aiming to capture significant improvements in objective understanding within a conversational setting. The results confirm the effectiveness of dialogue-based explanations over static ones in enhancing the understanding of machine learning model decisions among lay users (p < 0.004). Our experimental setup, encompassing three phases, effectively assessed users' objective understanding. We found that users engaging more deeply with specific questions about individual attributes showed the most significant gains in understanding. In contrast, static explanations resulted in lower model understanding, with participants often not changing their predictions after the learning phase. These results underscore the value of interactive explanations in XAI tools, fostering higher model understanding and engagement with explanations.

\begin{ack}
This research is funded by the Deutsche Forschungsgemeinschaft (DFG, German Research Foundation): TRR 318/1 Subproject B01, A dialog-based approach to explaining machine learning models, 2021 – 438445824.
\end{ack}

\bibliography{bibliography}
\end{document}